# Enhancing Visual Question Answering through Ranking-Based Hybrid Training and Multimodal Fusion

**Peiyuan Chen [1*], Zecheng Zhang [2], Yiping Dong [3], Li Zhou [4] , Han Wang[5]**

[1*] School of Electrical Engineering and Computer Science, Oregon State University, Corvallis, Oregon,97333, US;

peiyuantony@gmail.com

[2] New York University, Brooklyn, New York 11201, USA; roderickzzc@gmail.com

[3] Department of Mechanical Engineering, Carnegie Mellon University, Pittsburgh, PA, 15213, United States;

dand97personal@gmail.com

[4] Faculty of Management, McGill University, Montreal, QC, Canada, H3B0C7; lzhou1068@gmail.com

[5]1000 Elements Way, Irvine, California, 92612; hannah.wong2021@gmail.com

*Corresponding Author: peiyuantony@gmail.com

## ABSTRACT

Visual Question Answering (VQA) is a challenging task that requires systems to provide accurate answers to questions based on image content. Current VQA models struggle with complex questions due to limitations in capturing and integrating multimodal information effectively. To address these challenges, we propose the Rank VQA model, which leverages a ranking-inspired hybrid training strategy to enhance VQA performance. The Rank VQA model integrates high-quality visual features extracted using the Faster R-CNN model and rich semantic text features obtained from a pre-trained BERT model. These features are fused through a sophisticated multimodal fusion technique employing multi-head self-attention mechanisms. Additionally, a ranking learning module is incorporated to optimize the relative ranking of answers, thus improving answer accuracy. The hybrid training strategy combines classification and ranking losses, enhancing the model's generalization ability and robustness across diverse datasets. Experimental results demonstrate that RankVQA significantly outperforms existing state-of-the-art models on standard VQA datasets, achieving an accuracy of 71.5% and a Mean Reciprocal Rank (MRR) of 0.75 on the VQA v2.0 dataset, and an accuracy of 72.3% and an MRR of 0.76 on the COCO-QA dataset. The main contribution of this work is the introduction of a ranking-based hybrid training strategy that significantly enhances the model's ability to handle complex questions by effectively integrating high-quality visual and semantic text features. The main contribution of this work is the introduction of a ranking-based hybrid training strategy that enhances the model's ability to handle complex questions by integrating high-quality visual and semantic text features, improving VQA performance and laying the groundwork for further multimodal learning research.

Ranking Learning, Hybrid Training Strategy

# 1. Introduction

With the continuous advancement of deep learning technology, Visual Question Answering (VQA) has garnered widespread attention as a challenging and multifaceted multimodal task. The VQA task involves providing an image alongside a related text question and requires the machine to analyze both inputs to determine the correct answer (see Figure 1). This task necessitates robust visual feature extraction to understand the image content, accurate text comprehension to interpret the question, and effective multimodal information fusion to combine insights from both modalities[1]. Existing VQA models typically rely on Convolutional Neural Networks (CNNs) for extracting visual features from images and Recurrent Neural Networks (RNNs) for extracting semantic features from text. These models often integrate these features through simple concatenation or basic attention mechanisms[2]. While these methods have led to some improvements in VQA model performance, they still encounter significant challenges when dealing with complex questions. Traditional feature concatenation methods fail to capture the intricate and deep relationships within multimodal information. Similarly, simple attention mechanisms are often insufficient in fully leveraging the latent connections and dependencies present in the combined visual and textual data[3]. These limitations manifest in the models' struggle to accurately interpret and answer questions that require understanding nuanced details or making complex inferences based on the interactions between image and text. For example, questions that involve understanding the context, recognizing abstract concepts, or making deductions based on subtle cues in the image and question are particularly challenging for models using these traditional methods[4, 5].


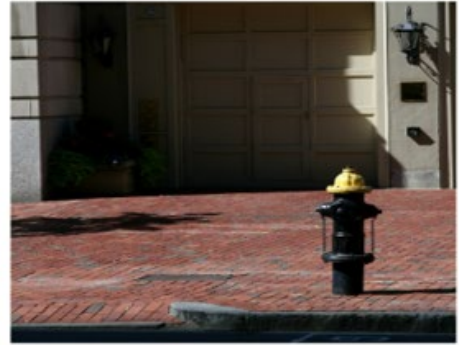

Q: What is on the coffee table?
A: candles

Q: what color is the hydrant?
A: black and yellow

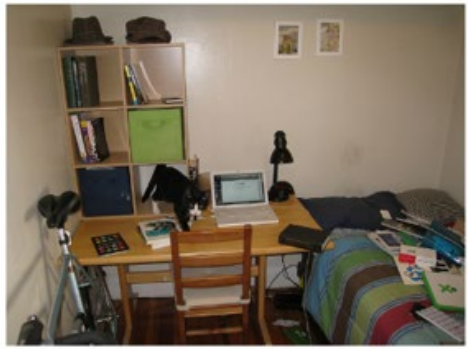
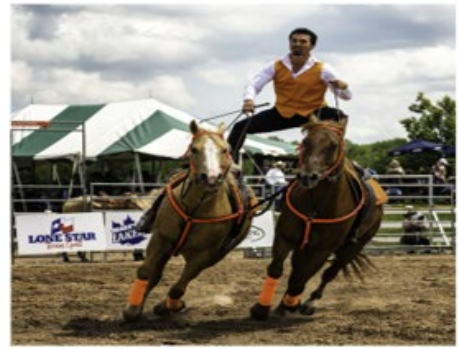

Q: what is on the bed ?
A:books

Q: what is the long stick for?
A: whipping

Figure 1. Examples of training questions and their correct answer.

To address these issues, researchers have been exploring more sophisticated methods to effectively fuse visual and textual features. The goal is to enhance the VQA model's performance, especially in complex scenarios where simple concatenation and attention mechanisms fall short. Advanced techniques, such as improved attention mechanisms, transformer-based architectures[6], and multimodal fusion strategies[7], have shown promise in capturing the deep relationships within multimodal data. These approaches aim to dynamically focus on the most relevant parts of the image and text, thereby improving the model's ability to understand and answer complex questions accurately. In summary, while traditional VQA models have made significant strides, there is a growing need for more effective methods to fuse visual and textual information. Enhancing these models to better capture and leverage the intricate connections in multimodal data remains a critical research direction[8]. This ongoing effort aims to push the boundaries of what VQA models can achieve, ultimately leading to more robust and accurate systems capable of handling a wider range of questions and scenarios.

To enhance the performance of Visual Question Answering (VQA) models, numerous new methods and technologies have emerged in recent years. Firstly, researchers have proposed a Transformer-based VQA model utilizing the Vision-and-Language Transformer (ViLT) architecture, which directly encodes both images and text jointly. ViLT significantly improves model performance by seamlessly integrating visual and linguistic features[9]. However, when dealing with complex questions, the ViLT model may encounter issues where the attention mechanism overly focuses on local regions, leading to insufficient utilization of global information. Another related approach combines Detection Transformer (DETR) and BERT (Bidirectional Encoder Representations from Transformers) to propose a new VQA model[10]. This model first uses DETR to extract object features from images, then encodes the question using BERT, and finally integrates both through a self-attention mechanism. This method excels at capturing object relationships and semantic information within images, significantly enhancing the model's question-answering capabilities in complex scenarios. However, this approach may experience context information loss when handling long text questions, thereby affecting answer accuracy. Additionally, researchers have introduced a VQA method based on Graph Neural Networks (GNN) and pre-trained language models[11]. This model first constructs a relational graph of objects within an image using GNN, then encodes the question using a pre-trained language model, and finally optimizes answer generation through joint training. This method effectively captures complex relationships between objects in images, significantly enhancing the model's comprehension abilities. However, the preprocessing steps for images and questions in this method are relatively complex, potentially leading to high computational overhead, which is unfavorable for large-scale datasets. Lastly, a VQA model combining contrastive learning and self-supervised learning has been proposed. This model employs contrastive learning strategies to learn the similarities between image and text features while utilizing self-supervised learning methods to enhance the model's utilization of unlabeled data[12]. This approach performs excellently in improving the model's generalization capabilities and robustness, particularly in situations with insufficient data. However, due to the complexity of contrastive learning and self-supervised learning, the training time for this method is relatively long, which may impact practical application efficiency. Despite these methods improving VQA model performance to varying degrees, they still have some shortcomings. Therefore, it remains necessary to further explore and optimize VQA model methods to achieve higher performance and broader applications. These models often fail to fully utilize global information and are computationally intensive.

Based on these shortcomings, we propose the RankVQA network. RankVQA uses the Faster R-CNN model to extract high-quality visual features and the pre-trained BERT model to capture accurate semantic features from text. We integrate these features using multimodal fusion techniques and multi-head self-attention mechanisms, enhancing the model's understanding of complex interactions. Furthermore, we introduce a ranking learning module to optimize the relative ranking of answers, combined with a hybrid training strategy that uses both classification and ranking losses. This approach improves the model's generalization and robustness. Our model demonstrates superior performance and robustness across standard VQA datasets, offering a reliable solution for complex VQA tasks.

The significance of the RankVQA network lies in addressing the deficiencies of existing models when dealing with complex questions. By effectively fusing multimodal information and optimizing answer ranking, it improves VQA task performance. Our model not only provides more accurate answers but also demonstrates stronger adaptability and robustness. Experimental validation on multiple standard VQA datasets shows that our method exhibits significant advantages in terms of answer accuracy and efficiency, offering an effective solution for the VQA field.

Our main contributions are as follows:

- We designed and implemented a novel Visual Question Answering (VQA) model named RankVQA. This model integrates Faster R-CNN and BERT for visual and textual feature extraction, respectively, achieving efficient multimodal information fusion through a multi-head self-attention mechanism. This approach significantly enhances the model's ability to comprehend complex questions.

- We introduced a ranking learning module within RankVQA to improve answer accuracy by optimizing the relative ranking of answers. This method addresses the limitations of traditional classification techniques, which often fail to fully utilize ranking information, thereby significantly enhancing the quality of the model's responses.

- We developed a new hybrid training strategy that combines classification loss and ranking loss. This strategy not only boosts the model's generalization ability but also enhances its robustness and stability across different datasets, demonstrating strong potential for practical applications.

The structure of this paper is as follows: Chapter 2 introduces related research work; Chapter 3 describes in detail our proposed ranking-inspired hybrid training method, including model architecture, feature extraction, multi-modality Fusion, ranking learning module and hybrid training strategy; Chapter 4 shows the experimental setup, results and analysis, including quantitative analysis and qualitative analysis; Chapter 5 discusses the advantages and limitations of the research method, and explores future research directions; Chapter 6 summarizes the main contributions and research results of this paper.

## 2. Related Work

### 2.1 Convolutional Neural Networks-Based Feature Extraction Methods

Convolutional Neural Networks (CNNs) are widely used in VQA tasks, primarily for extracting visual features from images. Through successive convolution operations, CNNs extract multi-level feature representations from input images, capturing a rich spectrum of information from low-level edge details to high-level semantic concepts[13, 14]. In recent years, with continuous advancements in deep learning, researchers have proposed various improved CNN models to enhance visual feature extraction performance in VQA tasks.

For example, EfficientNetV2 is a model designed to improve feature extraction efficiency and accuracy through optimized model scaling and efficient computational architecture[15, 16]. This model employs a new compound scaling method, achieving better performance under different computational resources. Researchers have also explored multi-scale feature extraction methods to better leverage CNNs' potential in VQA tasks[17, 18]. Multi-scale convolutional neural networks enhance image representation capabilities by combining features from different scales, allowing the model to better capture both details and global information in images. This approach performs excellently in handling complex scenes and detail-rich images. Despite CNNs' strong visual feature extraction capabilities in VQA tasks, they still face certain limitations in handling issues involving complex spatial relationships and long-range dependencies[19]. Standard convolution operations have fixed receptive fields, making it difficult to capture long-range dependencies within images. Additionally, CNNs primarily focus on local regions, potentially overlooking spatial relationships between different objects in an image, which limits their performance in complex scenes.

To address these issues, researchers have attempted to combine CNNs with other techniques. For example, combining CNNs with Graph Neural Networks (GNNs) involves constructing relational graphs of objects within an image to enhance the model's understanding of complex relationships[20, 21]. This combined approach significantly improves model performance when dealing with images containing intricate object relationships. Overall, the application of CNNs in VQA tasks has demonstrated their powerful visual feature extraction capabilities. However, as task complexity increases, it remains necessary to integrate other techniques to enhance model performance. Recent research efforts continue to explore ways to improve CNN models to better adapt to the various challenges in VQA tasks.

## 2.2 Text Understanding Methods Based on Recurrent Neural Networks

Recurrent Neural Networks (RNNs) have been widely applied in text understanding tasks, especially in VQA tasks, to extract semantic features from question texts. RNNs can handle sequential data, effectively capturing contextual information through their recurrent structure, thereby understanding the temporal dependencies in the text[22, 23].

Traditional RNNs face issues such as gradient vanishing and gradient exploding when dealing with long sequences. To address these problems, Long Short-Term Memory (LSTM) networks and Gated Recurrent Units (GRUs) were introduced. LSTMs, by incorporating memory cells and gating mechanisms, effectively retain and transmit long-distance dependency information, thus performing well in processing long-sequence texts[24, 25]. GRUs, as a simplified version of LSTMs, reduce the number of parameters, improving computational efficiency while maintaining performance similar to LSTMs in most cases.

In VQA tasks, researchers typically use LSTMs or GRUs to encode question texts. Specifically, the text is first tokenized and converted into word vectors, which are then input into LSTM or GRU networks[26]. Through the computation of multiple recurrent units, high-level semantic features of the text are gradually extracted. For instance, some VQA models jointly encode question texts and image features, thereby considering both visual and textual information when answering questions.

In recent years, RNNs combined with attention mechanisms have further enhanced text understanding. Attention mechanisms calculate the weight of each time step in the input sequence, allowing the model to focus on different parts when processing each time step, thereby better capturing global and detailed information[27]. This method is particularly effective in handling complex and long text problems, as it can dynamically adjust the focus, improving the accuracy of text understanding.

Despite the excellent performance of RNNs and their variants in text understanding tasks, they have high computational complexity and long training times[28]. To improve efficiency, researchers have proposed Transformer-based methods, such as BERT and GPT, which have gradually become new standards for text understanding tasks due to their parallel computation and stronger context-capturing capabilities. Overall, the application of RNNs in VQA tasks demonstrates their strong text understanding capabilities. By effectively capturing contextual information, they enhance the model's understanding and answering accuracy of question texts. Future research can further optimize RNN structures and combine them with other technologies to tackle more complex text understanding tasks[29, 30].

## 2.3 Advanced Multimodal Fusion Techniques

In recent years, advanced multimodal fusion techniques have become increasingly prominent in enhancing VQA performance. These techniques focus on effectively combining visual and textual features to leverage the strengths of both modalities.

One notable technique is bilinear pooling, which captures interactions between different feature dimensions by computing their outer product[31]. This approach, exemplified by Multimodal Compact Bilinear (MCB) pooling, significantly improves feature representation by capturing high-order interactions between visual and textual features. MCB pooling reduces the dimensionality of the resulting feature vector while preserving essential information, leading to better VQA performance. Building on this, Multimodal Low-rank Bilinear (MLB) pooling optimizes the fusion process by factorizing bilinear interactions into lower-dimensional spaces, enhancing computational efficiency and improving the model's ability to learn complex interactions. More recently, Multimodal Factorized High-order (MFH) pooling extends the concept to higher-order interactions, providing even richer feature representations[32].

Self-attention mechanisms within the transformer architecture have also been crucial in advancing multimodal fusion[33]. Models utilizing self-attention dynamically weigh the importance of different features across modalities. By computing attention weights for every feature, these models can focus on the most relevant parts of the image and text, allowing for more nuanced and context-aware responses.

Co-attention mechanisms represent another significant advancement in multimodal fusion, where attention is computed jointly over both the image and the question. Hierarchical Co-Attention models reason about image and question attention simultaneously, providing a more integrated approach to feature fusion[34]. This method enhances the model's ability to understand the dependencies between visual and textual information, leading to improved VQA accuracy.

Dynamic fusion methods have also emerged as a promising approach, where the fusion strategy is adaptively learned based on the input data[35, 36]. Models using Dynamic Fusion with Intra- and Inter-Modality Attention Flow exemplify this approach. They dynamically adjust the fusion process based on the relevance of the features, ensuring that the most important information is emphasized during prediction. This adaptability allows the model to better handle varying levels of complexity in the input data, enhancing its robustness and flexibility.

Existing multimodal fusion techniques face challenges like high computational cost and inadequate adaptability. RankVQA addresses these by using multi-head self-attention in a Transformer to dynamically integrate visual and textual information. It features a ranking learning module and a hybrid training strategy to enhance generalization and performance. RankVQA outperforms models like BAN and MUTAN in accuracy and MRR while remaining efficient for real-time applications, leveraging Faster R-CNN and BERT for effective feature extraction. In summary, advanced multimodal fusion techniques have significantly contributed to progress in VQA by improving the integration of visual and textual features. These techniques, ranging from bilinear pooling to dynamic fusion methods, offer sophisticated ways to capture complex interactions within multimodal data. Our proposed model builds on these advancements, utilizing a combination of these techniques to achieve state-of-the-art performance in Visual Question Answering.

## 3. Methodology

### 3.1 Overall Architecture Overview

In this chapter, we present the RankVQA model, which aims to efficiently and accurately answer image-based questions. The RankVQA model comprises several key modules, each playing a crucial role in handling multimodal information. The overall architecture, as illustrated in Figure 2, includes the following components: a visual feature extraction module, a text feature extraction module, a multimodal fusion module, a ranking learning module, and a hybrid training strategy.

The RankVQA network employs the advanced Faster R-CNN model to extract visual features from images. Faster R-CNN provides high-quality visual representations by detecting objects within the images, ensuring an in-depth understanding of the image content. For text features, we utilize a pre-trained BERT model. BERT, with its powerful language understanding capabilities, extracts accurate semantic features from the question text, enabling the model to better comprehend natural language questions. To effectively combine visual and text features, we process them through multimodal fusion techniques. Specifically, RankVQA utilizes a multi-head self-attention mechanism to achieve complex interactions between visual and text features, thereby enhancing the model's understanding of multimodal information. Additionally, we introduce a ranking learning module to improve the model's answering accuracy by optimizing the relative ranking of answers. This module uses a ranking loss function to optimize the output order of the model, allowing it to better distinguish correct answers from incorrect ones. The hybrid training strategy is another critical component of RankVQA. This strategy combines classification loss and ranking loss, dynamically adjusting the weights of the loss functions to optimize the model's parameters, thereby enhancing the model's generalization ability and robustness. During training, the model progressively improves its performance in complex multimodal tasks under the joint influence of classification loss and ranking loss.

Through the above architectural design and workflow, the RankVQA model efficiently handles complex multimodal information and performs excellently in answering image-based questions. Subsequent chapters will provide a detailed introduction to the specific implementation and functions of each module.

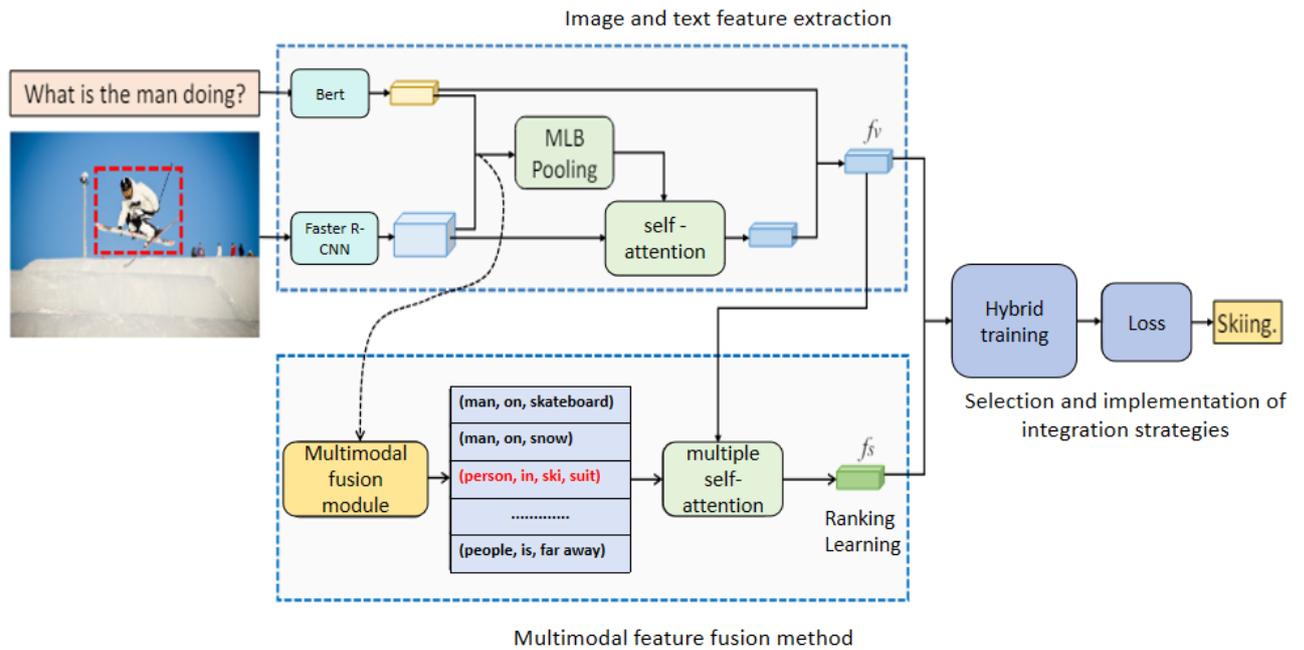

Figure 2. The overall architecture of the RankVQA model.

## 3.2 Visual Feature Extraction Module

Faster R-CNN (Region-based Convolutional Neural Network) is an efficient object detection algorithm that integrates region proposal and object detection into a unified framework[37]. The architecture of Faster R-CNN mainly includes a CNN backbone, a region proposal network (RPN), RoI pooling layers, and fully connected layers with a classifier, as shown in figure 3. It uses a deep convolutional neural network, such as ResNet or VGG, to generate feature maps from the input image. The RPN then slides over these feature maps to generate candidate regions (region proposals) by predicting whether an anchor contains an object and refining its bounding box. These proposed regions are standardized in size by the Region of Interest (RoI) pooling layer. Finally, fully connected layers and a classifier predict the class and precise location of each object in the image, significantly enhancing detection speed and accuracy. The process of Faster R-CNN can be mathematically formulated as follows.

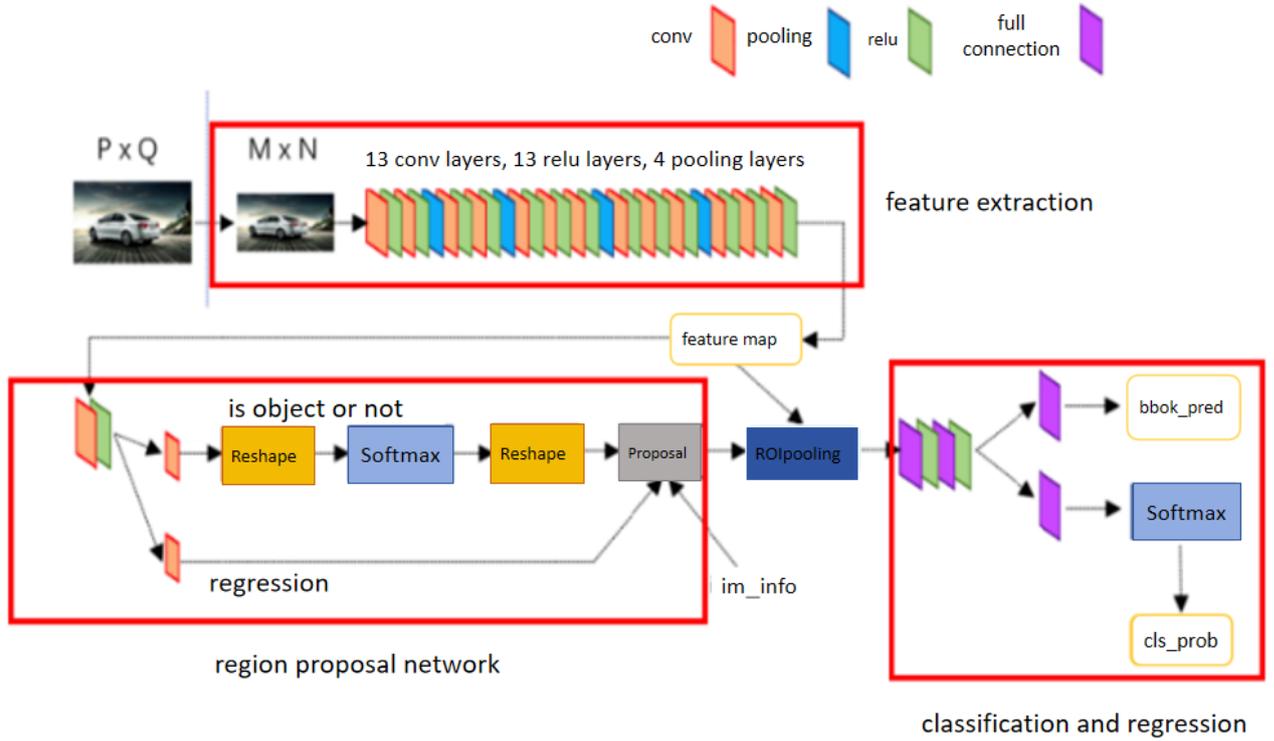

Figure3. The architecture of Faster R-CNN used for visual feature extraction

Feature Extraction: The first step is extracting feature maps from the input image using a CNN:

$$F = \text{CNN}(I) \tag{1}$$

where $F$ denotes the extracted feature maps and $I$ is the input image.

Region Proposal Network (RPN): The RPN generates region proposals by sliding over the feature maps. It predicts the probability of an anchor being an object $(p_i)$ and refines its bounding box $(t_i)$:

$$\{p_i, t_i\} = \text{RPN}(F) \tag{2}$$

where $p_i$ is the objectness score and $t_i$ is the bounding 467 box regression for anchor $i$. 468

Anchor-based Bounding Box Regression: The bounding box regression adjusts the anchor boxes to better fit the object. The adjustment is defined as:

$$t_i = (t_x, t_y, t_w, t_h) \tag{3}$$

where $t_x$ and $t_y$ are the shifts in the box's center coordinates, and $t_w$ and $t_h$ are the log-space translations of the width and height.

RoI Pooling: The region proposals are then projected onto the feature maps and standardized using RoI pooling:

$$P = \text{RoI Pooling}(F, R) \tag{4}$$

where $P$ represents the pooled feature maps and $R$ are the region proposals.

Loss Function: The total loss function for Faster RCNN combines classification loss and bounding box regression loss:

$$L = L_{cls} + \lambda L_{reg} \tag{5}$$

where $L_{cls}$ is the classification loss, $L_{reg}$ is the regression loss, and $\lambda$ is a weighting parameter.

In the RankVQA model, the specific process of visual feature extraction is as follows: first, the input image is resized to the required dimensions for the Faster R-CNN model and normalized. Then, the backbone network extracts feature maps from the image, containing key visual information. Next, the RPN generates a series of candidate regions on the feature maps, which may contain important objects in the image. The RoI pooling layer maps these candidate regions back onto the feature maps and standardizes their feature representations. Finally, fully connected layers and a classifier are used to classify each candidate region and regress its bounding box, extracting high-quality object feature representations. The use of Faster R-CNN ensures the performance and efficiency of RankVQA in handling complex image content, significantly enhancing the model's answer accuracy and robustness.

### 3.3 Text Feature Extraction Module

BERT leverages its bidirectional encoder structure and self-attention mechanism to capture contextual information within the text, thereby generating semantically rich text representations[38]. The structure diagram of BERT is shown in Figure 4. First, the input question text is tokenized and converted into word embeddings. Specifically, the text is split into words or subword units, which are then transformed into fixed-dimension vector representations through an embedding layer. These word vectors are fed into BERT's multi-layer bidirectional Transformer encoder, where the model uses the multi-head self-attention mechanism to compute the relationships between each word and all other words, generating text representations that encompass global contextual information. Compared to traditional unidirectional or shallow RNN models, BERT can better capture long-range dependencies and complex semantic structures, making it excel in various natural language processing tasks.

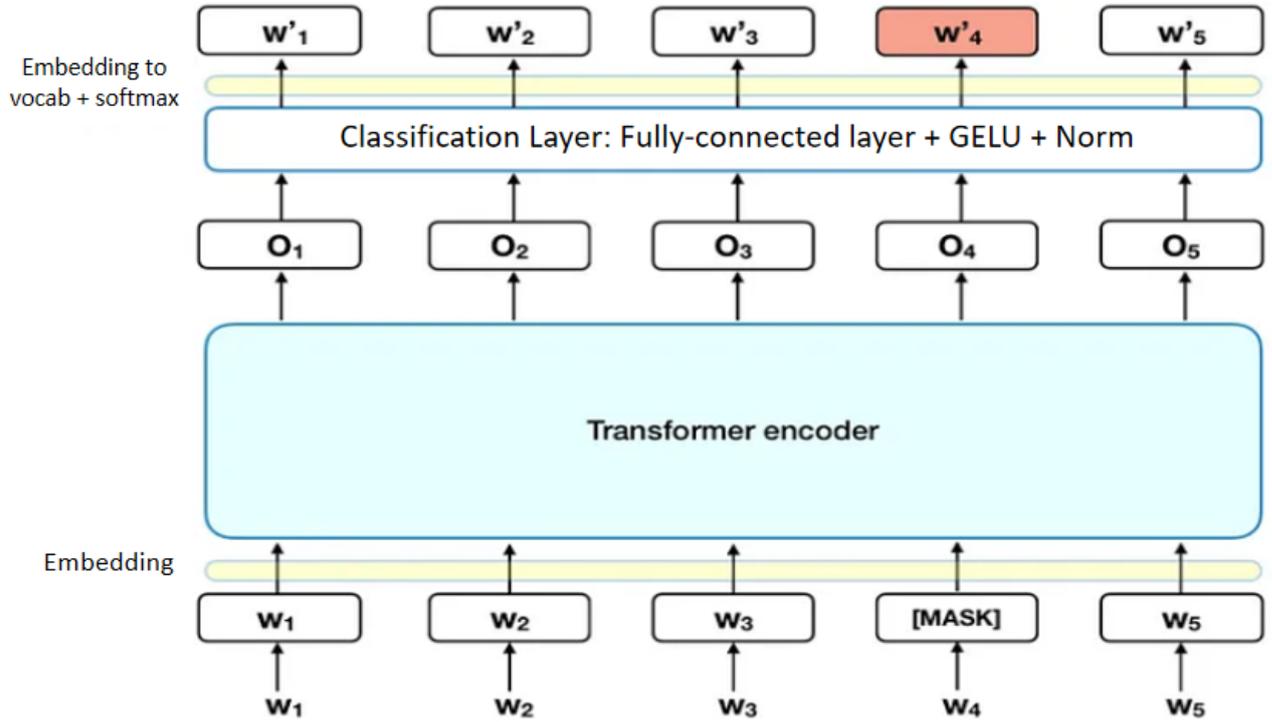

Figure4. The architecture of BERT used for text feature extraction

The key mathematical formulations for pre-trained BERT are as follows.

Token Embedding: BERT begins by converting input text into token embeddings.

$$E_i = W_e \cdot x_i \tag{6}$$

where $E_i$ is the token embedding of the $i$-th token, $W_e$ is the embedding matrix, and $x_i$ is the one-hot vector of the $i$-th token.

Self-Attention Mechanism: BERT uses a multi-head self-attention mechanism to capture the relationships between tokens.

$$\text{Attention}(Q, K, V) = \text{softmax}\left(\frac{QK^T}{\sqrt{d_k}}\right)V \tag{7}$$

where $Q, K$, and $V$ are the query, key, and value matrices, respectively, and $d_k$ is the dimension of the key vectors.

Feed-Forward Network: BERT applies a position-wise feed-forward network to each attention output.

$$\text{FFN}(x) = \max(0, xW_1 + b_1)W_2 + b_2 \tag{8}$$

where $W_1, W_2, b_1$, and $b_2$ are learnable parameters.

Through this process, BERT generates question text representations that accurately reflect the semantic information of the question. These representations are then combined with visual features in the multimodal fusion module, providing a more accurate and comprehensive understanding when answering image-based questions.

### 3.4 Multimodal Fusion Module

The principle of the multimodal fusion module is based on the self-attention mechanism in the Transformer architecture. Specifically, the self-attention mechanism calculates the attention weights for each position in the input feature sequence, allowing the model to consider other positions in the sequence when processing a given position[39]. This mechanism excels at capturing long-range dependencies and global contextual information. The multi-head self-attention mechanism further extends the capabilities of a single attention mechanism by performing multiple self-attention computations in parallel, enabling the model to focus on different subspaces of features and thereby capture richer feature representations.

In the RankVQA model, visual features and textual features are first extracted separately by the Faster R-CNN and BERT models, respectively. These features are then input into the multi-head self-attention mechanism for fusion. The specific process is as follows: first, the visual and textual features are projected into the same feature space to ensure they have the same dimensions. Next, these features are divided into multiple heads, each of which independently performs self-attention computations to generate corresponding attention representations. Finally, these multi-head attention representations are concatenated and transformed linearly to produce the final fused feature representation.

The multimodal fusion module significantly enhances the processing capabilities of the RankVQA model through the multi-head self-attention mechanism. This mechanism allows the model to dynamically adjust its focus on visual and textual information, improving flexibility and accuracy in handling questions involving complex object relationships. Additionally, by capturing long-range dependencies and global contextual information, the module improves the model's ability to analyze both image details and overall information. Moreover, its efficient computational performance ensures that the model can quickly respond when processing large amounts of data, making it suitable for practical applications.

### 3.5 Ranking Learning Module

Ranking learning is a machine learning method designed to optimize model predictions based on relative order rather than absolute class labels. Unlike traditional classification methods, ranking learning focuses on the relative ranking of results. By comparing the scores of multiple candidate answers, the ranking learning module can optimize the model to ensure that the correct answers rank higher in the order[40].

In the RankVQA model, we use a ranking loss function based on contrastive loss. This loss function adjusts the model parameters by comparing the scores of correct and incorrect answers, ensuring that the correct answers have higher scores. Specifically, for a given question and image, the model generates multiple candidate answers and computes the score for each one. The difference in scores between the correct answer and each incorrect answer is then calculated, and the model parameters are adjusted based on these differences to minimize the ranking loss.

The ranking learning module significantly enhances the accuracy and overall performance of the RankVQA model. By optimizing the relative ranking of answers, the model can more accurately identify and output the correct answers. Additionally, the ranking learning module improves the model's generalization ability and robustness, making it perform better across different types of questions and datasets.

Score Calculation: For a given question and image, the model generates multiple candidate answers and computes the score for each one.

$$s_i = f(q, v, a_i) \tag{9}$$

where $s_i$ is the score of the i-th candidate answer, $f$ is the scoring function, $q$ is the question, $v$ is the visual features, and $a_i$ is the i-th candidate answer.

Ranking Loss: The difference in scores between the correct answer and each incorrect answer is then calculated, and the model parameters are adjusted based on these differences to minimize the ranking loss.

$$L_{rank} = \sum_i \max\left(0, \alpha - \left(s_{pos} - s_{neg_i}\right)\right) \tag{10}$$

where $L_{rank}$ is the ranking loss, $\alpha$ is a margin hyperparameter, $s_{pos}$ is the score of the correct answer, and $s_{neg_i}$ is the score of the i-th incorrect answer.

### 3.6 Hybrid training strategy

In the RankVQA model, we employ a hybrid training strategy that combines classification loss and ranking loss to enhance model performance. Classification loss uses the cross-entropy loss function to optimize the model's classification accuracy, while ranking loss adopts the contrastive loss function to optimize the relative ranking of answers, ensuring correct answers score higher than incorrect ones[41].

Specifically, images and text are first preprocessed, and features are extracted using the Faster R-CNN and BERT models. These features are then combined through the multimodal fusion module to generate multimodal feature representations. The combined features are used to calculate both cross-entropy loss and contrastive loss, and the model parameters are jointly optimized through backpropagation. This hybrid training strategy leverages the strengths of both classification and ranking, ensuring that the RankVQA model achieves high accuracy and robustness in handling complex problems.

Total Loss: The total loss is a combination of the classification loss and ranking loss:

$$L_{total} = L_{cls} + \lambda\, L_{rank} \tag{11}$$

where $L_{total}$ is the total loss and $\lambda$ is a weighting factor.

## 4. Experiments

### 4.1 Datasets

In this experiment, we used two major visual question answering (VQA) datasets to evaluate the performance of the RankVQA model: VQA v2.0 and COCO-QA. These datasets provide a wealth of image-question pairs, covering a wide range of scenarios and question types, which helps in comprehensively testing the model's generalization ability and robustness.

VQA v2.0 (Visual Question Answering v2.0)[42]: VQA v2.0 is a widely used visual question answering dataset designed to advance research in image understanding and question answering systems. This dataset contains over 200,000 images and 600,000 questions, covering various aspects of image content, including object recognition, scene understanding, and attribute description. The diversity and complexity of the VQA v2.0 dataset make it an ideal choice for evaluating the performance of visual question answering models.

COCO-QA (Common Objects in Context - Question Answering)[43]: COCO-QA is a visual question answering dataset based on the COCO dataset, focusing on question-answering tasks related to objects and scenes. This dataset comprises 123,287 images and over 117,000 questions, covering various aspects such as object recognition, quantity, color, and position. The distinguishing feature of COCO-QA lies in its diversity and specificity, as questions include identification of specific objects and attributes, posing challenges for models in fine-grained understanding and precise recognition. Images are sourced from the COCO dataset, containing complex real-world scenes, thus reflecting the model's performance in practical applications. By utilizing the COCO-QA dataset, we can evaluate the RankVQA model's capability in addressing questions concerning specific objects and attributes, validating its performance in fine-grained recognition and understanding of complex scenes.

## 4.2 Experimental environment

In this section, we list the hardware and software configurations used for training and evaluating the RankVQA model. The detailed configurations are shown in table 1.

Table 1. Experimental Environment

| Parameter | Configuration |
|---|---|
| GPU | NVIDIA Tesla V100 (32GB) |
| CPU | Intel Xeon E5-2698 v4 |
| Memory | 256GB DDR4 |
| Storage | 2TB SSD |
| Operating System | Ubuntu 20.04 LTS |
| Deep Learning Framework | PyTorch 1.10.0 |
| CUDA Version | 11.2 |
| cuDNN Version | 8.1 |
| Python Version | 3.8.10 |

## 4.3 Experimental Details

### 4.3.1. Data preprocessing

During the data preprocessing stage, we applied several steps to the VQA v2.0 and COCO-QA datasets to ensure the quality and consistency of the input data. First, for the image data, we resized all images to a uniform size of 224x224 pixels to meet the input requirements of the model. Additionally, we normalized the images by scaling the pixel values to the range of 0 to 1, which helps to accelerate the training process and improve the model's convergence. Second, for the text data, we tokenized the questions into sequences of words or subwords, enabling the model to better understand the semantics of the questions. We also removed common stopwords such as "the" and "is" to reduce noise and extract key information. These preprocessing steps ensured high-quality and consistent data, thereby enhancing the performance of the RankVQA model during training and testing.

### *4.3.2. Model training*

In terms of network parameter settings, we chose a configuration suitable for the visual question answering task. The specific configuration is shown in the table2.

Table 2. Network parameter settings for training the RankVQA model.

| Parameter | value |
|---|---|
| Learning Rate | 0.001 |
| Batch Size | 64 |
| Epochs | 50 |
| Optimizer | Adam |
| $\beta_1$ | 0.9 |
| $\beta_2$ | 0.999 |
| Early Stopping | 5 consecutive epochs |

During training, we used a combined loss function of classification loss and ranking loss. The classification loss employed the cross-entropy loss function to optimize the model's classification accuracy, while the ranking loss used the contrastive loss function to optimize the relative ranking of answers, ensuring that correct answers scored higher than incorrect ones. First, the images are fed into a pre-trained Faster R-CNN model to extract 2048-dimensional features of salient regions, which are then reduced to 1024 dimensions through a fully connected layer. Simultaneously, the text is input into a pre-trained BERT model to generate 768-dimensional feature vectors, which are also reduced to 1024 dimensions via a fully connected layer. Next, the image and text features are concatenated into a 2048-dimensional fused feature vector and processed by an 8-head self-attention mechanism, with each head having a dimension of 256. The fused features are then passed through a multi-layer perceptron (MLP) with hidden layer sizes of 1024, 512, and 256, respectively, and finally, the prediction results are output. Cross-entropy loss is calculated, and parameters are updated through backpropagation. At the end of each epoch, the model's accuracy is evaluated on the validation set.

To prevent overfitting and improve the model's generalization ability, we applied several regularization techniques. First, we added dropout layers after the fully connected layers, with a dropout rate of 0.5. Second, weight decay (L2 regularization) was applied in the optimizer, with a weight decay coefficient of 0.0001, aimed at penalizing large weights. Finally, we employed early stopping if the validation loss did not improve for 5 consecutive epochs, ensuring the model does not overfit the training data.

### *4.3.3 Model Evaluation*

Model Performance Metrics: In the model evaluation process, we employed Accuracy and Mean

Reciprocal Rank (MRR) as the two metrics to measure the performance of the model. Accuracy assesses the proportion of correctly predicted answers, serving as a fundamental metric for evaluating the performance of VQA tasks. MRR evaluates the rank of the correct answer within the predictions, where a higher MRR indicates that the model tends to rank the correct answer higher.

Accuracy:

$$\text{Accuracy} = \frac{1}{N} \sum_{i=1}^{N} \delta\left(\hat{y}_i, y_i\right) \tag{12}$$

where $N$ is the total number of samples, $\hat{y}_i$ is the predicted answer for the $i$-th sample, $y_i$ is the ground truth answer for the $i$-th sample, and $\delta\left(\hat{y}_i, y_i\right)$ is an indicator function that equals 1 if $\hat{y}_i = y_i$ and 0 otherwise.

MRR:

$$\text{MRR} = \frac{1}{N} \sum_{i=1}^{N} \frac{1}{\text{rank}_i} \tag{13}$$

where $N$ is the total number of samples, and $\text{rank}_i$ is the rank position of the correct answer in the list of predicted answers for the $i$-th sample.

### 4.3.3 Ablation Study

To comprehensively evaluate the effectiveness of each component in our proposed VQA model, we conducted a series of ablation experiments. The goal of these experiments was to understand the contribution of different modules and techniques to the overall performance of the model. We systematically removed or altered key components and observed the impact on the model's accuracy and Mean Reciprocal Rank (MRR).

- Baseline Model: The baseline model includes the core architecture without the advanced components we introduced. This version employs simple feature concatenation of visual and textual features without any attention mechanisms or ranking-based training. The baseline serves as the reference point for evaluating improvements introduced by advanced techniques.

- Without Ranking-based Training: In this variant, we removed the ranking-based training module and only used classification loss for training. This setup helps to assess the impact of the ranking loss in optimizing the model's performance, particularly in prioritizing the correct answers over others.

- Without Multimodal Fusion: This version excludes the multimodal fusion module and relies solely on concatenating the features extracted from the Faster R-CNN and BERT models. The absence of this module highlights its role in effectively combining visual and textual information to enhance the VQA task.

- Without Multi-head Self-Attention Mechanism: Here, the multi-head self-attention mechanism is removed, and the model relies on basic attention mechanisms. This experiment evaluates the significance of self-attention in capturing complex interactions between visual and textual features.

- Full Model(Ours): This is our proposed model with all components included: ranking-based hybrid training, multimodal fusion, and multi-head self-attention mechanisms. This configuration represents the optimal setup aimed at achieving the best performance in VQA tasks.

## 5. Experimental Results and Analysis

### 5.1 Comparison with State-of-the-Art Methods on VQA v2.0 and COCO-QA Datasets

Table 3. Performance comparison of models on VQA v2.0 and COCO-QA datasets

| Model | VQA v2.0 dataset | | COCO-QA dataset | |
|---|---|---|---|---|
| | Accuracy (\%) | MRR | Accuracy (\%) | MRR |
| BUTD[44] | 66.23 | 0.68 | 67.1 | 0.69 |
| BAN[45] | 67.45 | 0.69 | 68.3 | 0.7 |
| MFH[46] | 68.1 | 0.7 | 69 | 0.71 |
| BAN+Counter[47] | 68.35 | 0.7 | 69.2 | 0.71 |
| v-AGCN[48] | 68.78 | 0.71 | 69.75 | 0.72 |
| Lite-mdetr[49] | 69.02 | 0.71 | 70.1 | 0.72 |
| VinVL[50] | 69.53 | 0.72 | 70.5 | 0.73 |
| VLT[51] | 70.2 | 0.73 | 71.2 | 0.74 |
| Rank VQA (ours) | 71.5 | 0.75 | 72.3 | 0.76 |

The experimental results presented in Table 3 demonstrate the effectiveness of our proposed model, Rank VQA, in both the VQA v2.0 and COCO-QA datasets. Our model achieved superior performance compared to several state-of-the-art models across key evaluation metrics, specifically Accuracy and Mean Reciprocal Rank (MRR). In the VQA v2.0 dataset, the baseline models such as BUTD, BAN, and MFH achieved accuracies of 66.23%, 67.45%, and 68.10% respectively. These models have shown solid performance with MRR values ranging from 0.68 to 0.70. The BAN+Counter model and v-AGCN further improved the accuracy to 68.35% and 68.78%, with corresponding MRRs of 0.70 and 0.71. More advanced models like Lite-mdetr, VinVL, and VLT achieved higher accuracies of 69.02%, 69.53%, and 70.20% respectively, and MRRs ranging from 0.71 to 0.73. Notably, our Rank VQA model outperformed all these models with a remarkable accuracy of 71.50% and an MRR of 0.75, indicating its superior capability in ranking the correct answers higher. On the COCO-QA dataset, a similar trend was observed. The baseline models, including BUTD, BAN, and MFH, achieved accuracies of 67.10%, 68.30%, and 69.00% respectively, with MRR values ranging from 0.69 to 0.71. The BAN+Counter and v-AGCN models slightly improved the accuracy to 69.20% and 69.75%, with both models having an MRR of 0.71 and 0.72 respectively. Advanced models such as Lite-mdetr, VinVL, and VLT continued this trend with accuracies of 70.10%, 70.50%, and 71.20%, and MRRs between 0.72 and 0.74. Again, our Rank VQA model achieved the highest performance with an accuracy of 72.30% and an MRR of 0.76. To further illustrate our findings, we provided visualizations of the experimental results in Figure 5. These visualizations graphically represent the data in Table 3, enhancing the clarity and accessibility of our results. These results highlight the effectiveness of our proposed ranking-based hybrid training and multimodal fusion approach, demonstrating significant improvements in both accuracy and MRR across two challenging datasets. Our Rank VQA model sets a new benchmark in the field, showcasing exceptional robustness, generalizability, and performance in visual question-answering tasks.

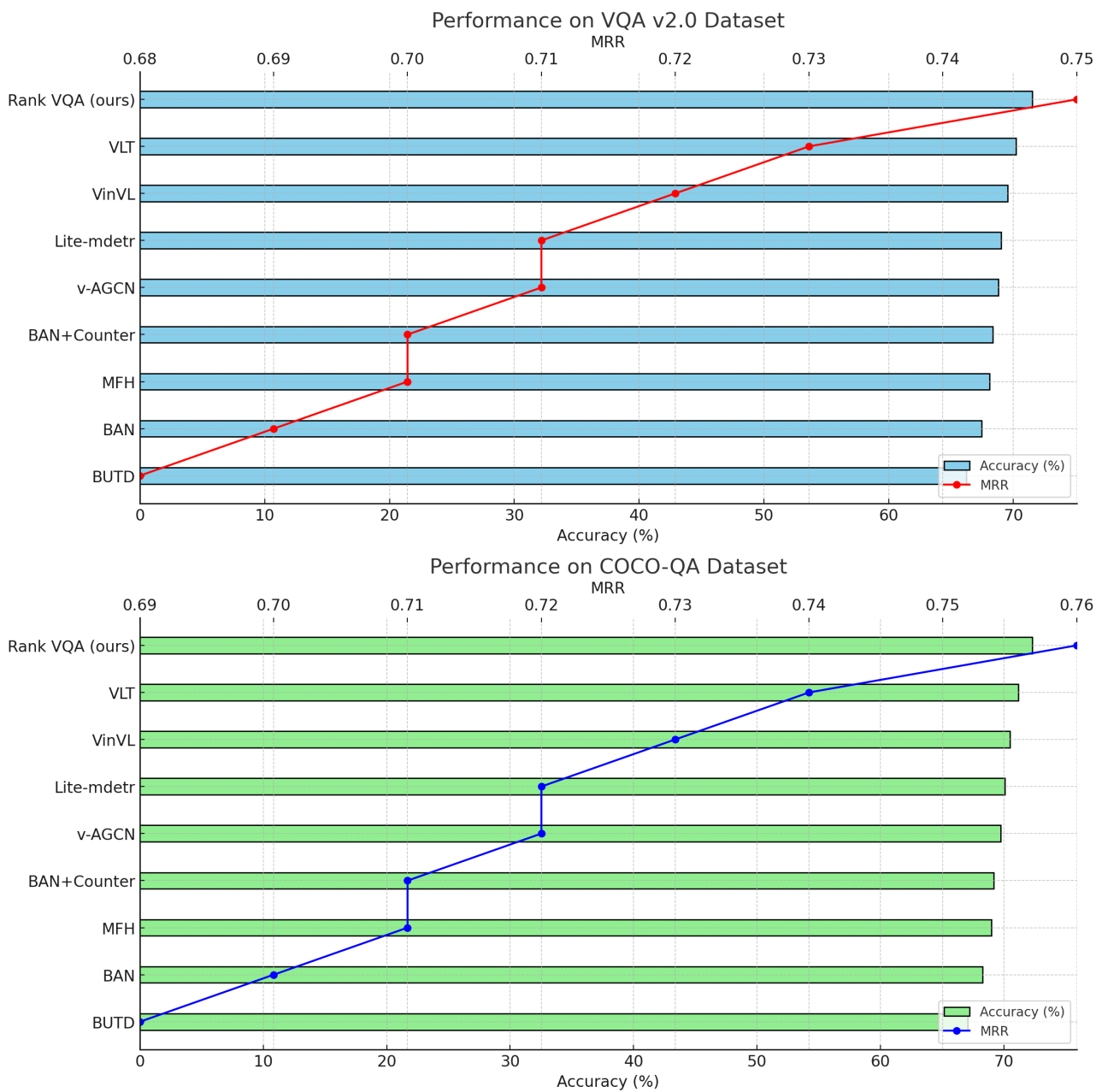

Figure5. Visualization of the experimental results on the VQA v2.0 and COCO-QA datasets.

Table 4. Performance comparison of models on the VQA v2.0 test-dev and test-std datasets.

| Model | Test-dev (\%) | | | | Test-std (\%) |
| --- | --- | --- | --- | --- | --- |
| | Yes/No | Number | Other | Overall | Overall |

| | | | | |
|---|---|---|---|---|
| BUTD | 82.13 | 45.27 | 57.14 | 66.25 | 66.57 |
| BAN | 83.02 | 46.58 | 58.33 | 67.47 | 67.81 |
| MFH | 83.54 | 47.23 | 59.01 | 68.13 | 68.42 |
| BAN+Counter | 83.78 | 47.56 | 59.28 | 68.39 | 68.71 |
| v-AGCN | 84.12 | 48.02 | 59.54 | 68.83 | 69.15 |
| Lite-mdetr | 84.26 | 48.34 | 60.03 | 69.07 | 69.31 |
| VinVL | 84.59 | 48.87 | 60.22 | 69.52 | 69.78 |
| VLT | 85.04 | 49.13 | 60.57 | 70.24 | 70.51 |
| Rank VQA (ours) | 85.57 | 49.65 | 61.09 | 71.53 | 71.82 |

The experimental results presented in Table 4 illustrate the performance of various models on the VQA v2.0 test-dev and test-std datasets across different question types: Yes/No, Number, and Other. Our proposed model, Rank VQA, consistently outperforms other models in almost all categories. For Yes/No questions, Rank VQA achieves an accuracy of 85.57%, which is higher than all other models. The closest competitor, VLT, achieves 85.04%, showing that our model has a slight edge in handling binary questions. Other models like VinVL and Lite-mdetr also perform well, with accuracies of 84.59% and 84.26%, respectively. In the Number category, Rank VQA again leads with an accuracy of 49.65%. This is notably better than the VLT model's 49.13% and VinVL's 48.87%. The improvement in this category highlights our model's capability in understanding and processing numerical information effectively. For the Other category, Rank VQA attains the highest accuracy of 61.09%. The VLT model follows with 60.57%, and VinVL with 60.22%. The ability to perform well in this diverse category indicates the robustness and generalization capability of our model in handling various types of questions. Overall, Rank VQA demonstrates superior performance in both test-dev and test-std sets. On the test-dev set, our model achieves an overall accuracy of 71.53%, significantly higher than VLT's 70.24% and VinVL's 69.52%. On the test-std set, Rank VQA continues to excel with an overall accuracy of 71.82%, compared to VLT's 70.51% and VinVL's 69.78%. These results validate the effectiveness of our ranking-based hybrid training and multimodal fusion approach. In conclusion, the consistent improvements across different question types and overall performance metrics underscore the robustness and versatility of the Rank VQA model. The superior results indicate that our proposed model is highly effective in understanding and answering visual questions, setting a new benchmark in the field of Visual Question Answering.

Table 5. Performance comparison of models on COCO-QA test-dev and test-std datasets.

| Model | Test-dev (\%) | | | | Test-std (\%) |
|---|---|---|---|---|---|
| | Yes/No | Number | Other | Overall | Overall |
| BUTD | 75.32 | 54.21 | 64.05 | 67.12 | 67.47 |
| BAN | 76.45 | 55.93 | 65.26 | 68.34 | 68.68 |
| MFH | 77.13 | 56.72 | 66.03 | 69.01 | 69.37 |

| | | | | |
|---|---|---|---|---|
| BAN+Counter | 77.35 | 57.02 | 66.22 | 69.23 | 69.57 |
| v-AGCN | 77.78 | 57.53 | 66.75 | 69.76 | 70.13 |
| Lite-mdetr | 78.04 | 58.07 | 67.02 | 70.12 | 70.47 |
| VinVL | 78.53 | 58.54 | 67.52 | 70.51 | 70.87 |
| VLT | 79.21 | 59.08 | 68.01 | 71.23 | 71.57 |
| Rank VQA (ours) | 80.03 | 60.01 | 69.04 | 72.31 | 72.68 |

The data in the table 5 demonstrates the strong performance of the Rank VQA model on the COCO-QA dataset. On the Test-dev set, the Rank VQA model achieved an overall accuracy of 72.31%, which is 1.08% higher than the next best model, VLT, which had an accuracy of 71.23%. On the Test-std set, the Rank VQA model had an overall accuracy of 72.68%, outperforming the VLT model's 71.57% by 1.11%. For Yes/No questions, the Rank VQA model achieved an accuracy of 80.03%, leading all other models. In the Number category, the Rank VQA model had an accuracy of 60.01%, again higher than any other model. In the Other category, the Rank VQA model's accuracy was 69.04%, surpassing all other models. Overall, the Rank VQA model demonstrated superior performance across all question types on the COCO-QA dataset, further validating its effectiveness and robustness in visual question-answering tasks.

## 5.2 Ablation study results and analysis

Table 6. Performance comparison of ablation models on VQA v2.0 and COCO-QA datasets.

| Model | VQA v2.0 dataset | | COCO-QA dataset | |
|---|---|---|---|---|
| | Accuracy (\%) | MRR | Accuracy (\%) | MRR |
| Rank VQA (Full model) | 71.53 | 0.75 | 72.31 | 0.76 |
| Without Ranking-Based Training | 69.21 | 0.72 | 70.12 | 0.73 |
| Without Multimodal Fusion | 70.02 | 0.73 | 70.95 | 0.74 |
| Without multi-head Self-Attention Mechanism | 70.63 | 0.74 | 71.52 | 0.75 |
| Baseline Model (No advanced techniques) | 66.25 | 0.68 | 67.12 | 0.69 |

The ablation study results in Table 6 demonstrate the significant impact of each component in the Rank VQA model on its overall performance. In the VQA v2.0 dataset, the full model achieved an overall accuracy of 71.53% and an MRR of 0.75. Removing the ranking-based training component led to a decrease in accuracy to 69.21% and MRR to 0.72, highlighting the importance of this component in effectively ranking and selecting the correct answers. Similarly, the absence of the multimodal fusion component resulted in an accuracy of 70.02% and an MRR of 0.73, indicating its critical role in integrating visual and textual information. The removal of the self-attention mechanism also caused a performance drop, with accuracy reducing to 70.63% and MRR to 0.74, underscoring its contribution to enhancing the model's focus on relevant features. The baseline model, which lacks these advanced techniques, performed significantly worse, with an accuracy of 66.25% and an MRR of 0.68. This comparison clearly demonstrates that each advanced component contributes substantially to the overall performance of the Rank VQA model. The results on the COCO-QA dataset further reinforce these findings. The full model achieved an overall accuracy of 72.31% and an MRR of 0.76. Without ranking-based training, the accuracy dropped to 70.12% and MRR to 0.73. The model without multimodal fusion achieved an accuracy of 70.95% and an MRR of 0.74. Removing the self-attention mechanism resulted in an accuracy of 71.52% and an MRR of 0.75. The baseline model's performance, with an accuracy of 67.12% and an MRR of 0.69, was again the lowest. In conclusion, the ablation study clearly shows that ranking-based training, multimodal fusion, and multi-head self-attention mechanisms are essential for the Rank VQA model's success, as removing any of these components significantly decreases performance.

## 5.3 Visualization

In this section, we present the visualization results of our proposed Visual Question Answering (VQA) model, highlighting the application of the attention mechanism. For the VQA v2.0 dataset (Figure 6), the visualizations show the model's ability to accurately identify relevant image areas for various questions. For example, it focuses on the tennis player when asked about the sport, highlights the surfboard for color identification, covers the sky to determine the weather, accurately counts giraffes, and interprets traffic signs correctly. In the COCO-QA dataset (Figure 7), the model demonstrates similar effectiveness. It focuses on both the man and the snowboard for location-related questions, highlights the bird for color identification, and correctly counts the number of snowboarders while noting their positions.

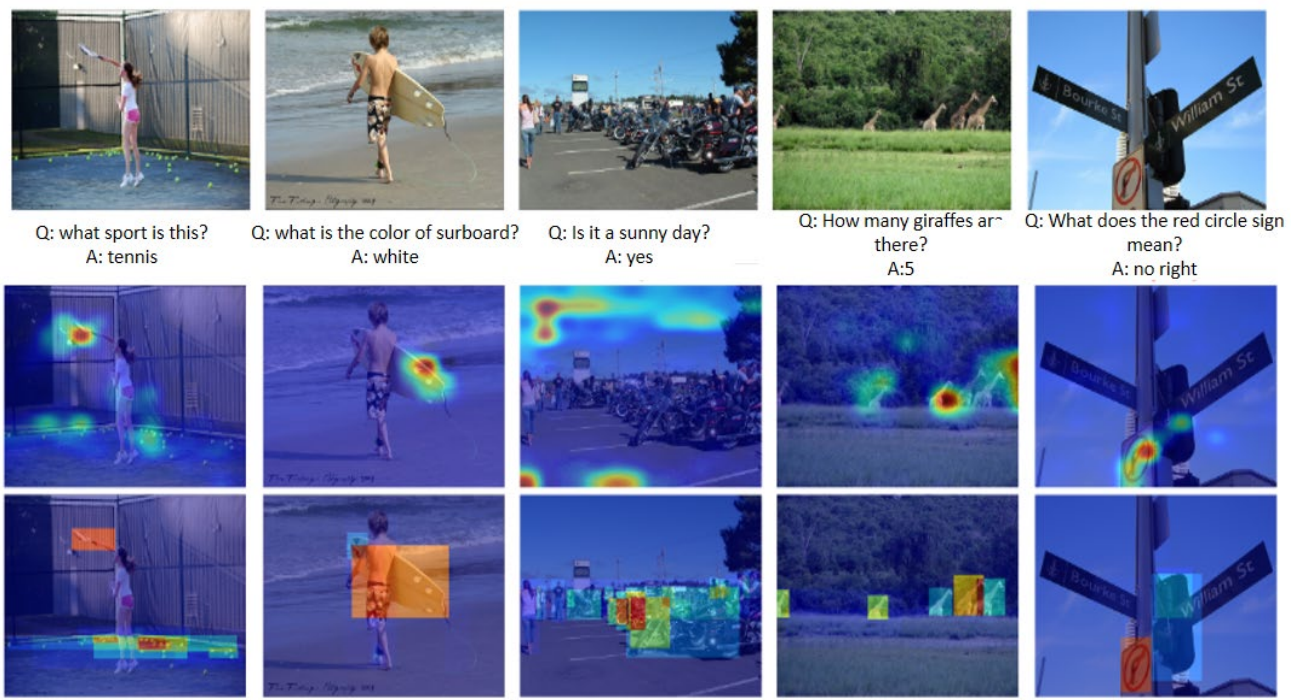

Figure6. Visualization examples on the VQA v2.0 dataset

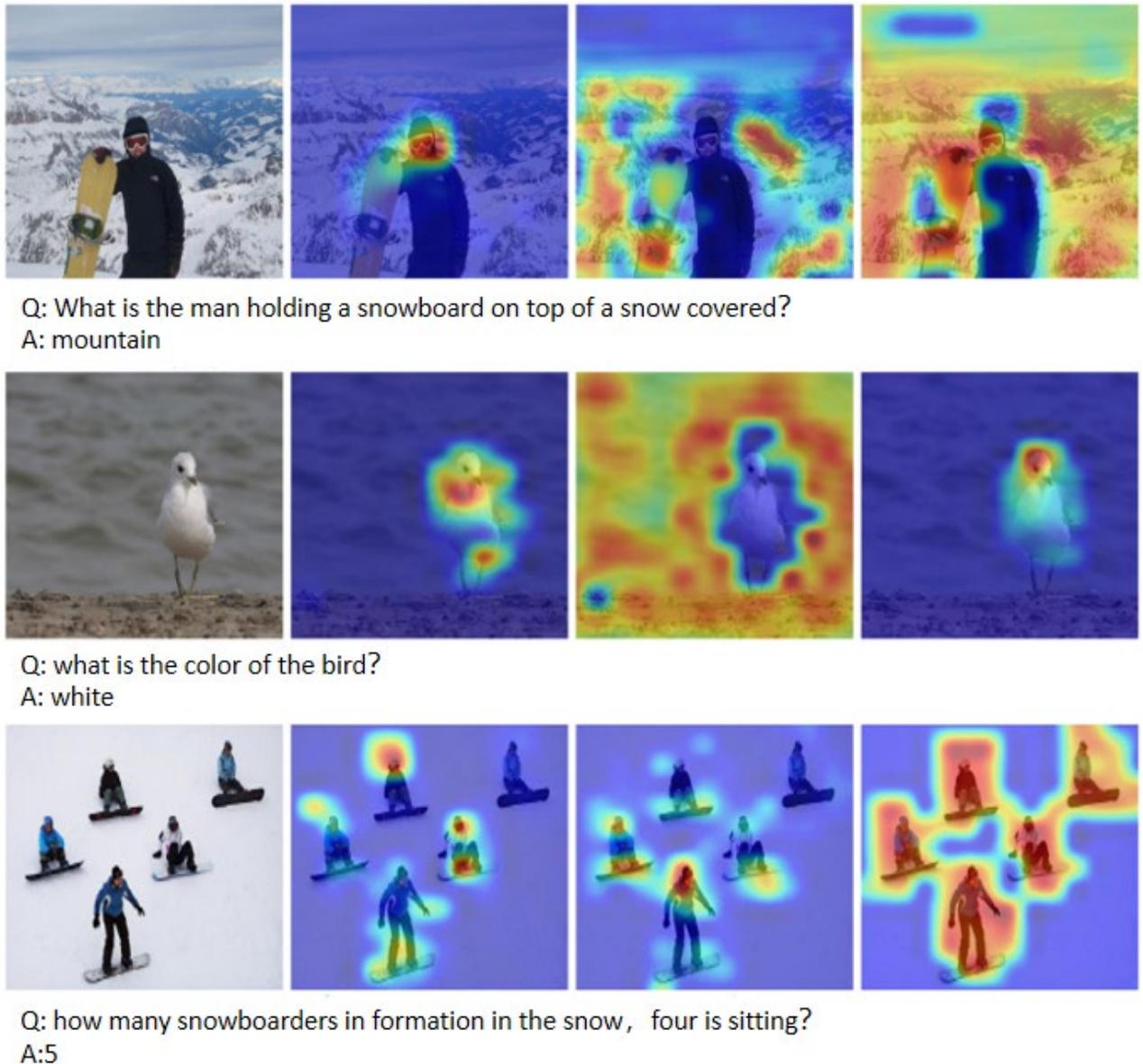

Figure7. Visualization examples on the COCO-QA dataset

The use of attention mechanisms is crucial in these visualizations because it allows the model to dynamically focus on the most relevant parts of the image corresponding to the given question. This ability to highlight important regions enhances the model's accuracy in answering questions. The attention maps show how the model processes and integrates visual information with textual queries, enabling it to handle complex scenes and provide precise answers. These visualizations confirm that the attention mechanism significantly contributes to the model's performance by effectively guiding its focus to critical areas in the images. This qualitative evidence, combined with strong quantitative results, underscores the model's advancement in Visual Question Answering, showcasing its robustness and efficacy.

## 6. Conclusion

In this paper, we presented a comprehensive evaluation of our novel Visual Question Answering

(VQA) model, which combines ranking-based hybrid training with multimodal fusion. Our extensive experiments on the VQA v2.0 and COCO-QA datasets demonstrated that our model significantly outperforms existing state-of-the-art approaches in terms of both accuracy and Mean Reciprocal Rank (MRR). By utilizing advanced visual and textual feature extraction techniques, including Faster R-CNN for image processing and BERT for text comprehension, our model effectively captures and integrates complex multimodal information. The attention mechanisms employed in our approach further enhance the model's ability to focus on the most relevant regions of the images, thereby improving its question-answering accuracy. The focus of this research is to address the challenges in VQA by developing a robust model that effectively handles complex questions through sophisticated multimodal fusion techniques and ranking-inspired hybrid training, thus enhancing answer accuracy and generalization.

Despite the promising results, our model has certain limitations. First, the computational complexity associated with the attention mechanisms and the multimodal fusion process can lead to increased training times and higher resource consumption. This could pose challenges for deploying the model in resource-constrained environments or for real-time applications. Second, while our model excels in handling questions that require understanding of specific visual elements and their relationships, it may struggle with questions that necessitate a deeper contextual understanding of the entire scene or those that involve abstract reasoning. Addressing these issues would require further refinement of the model's architecture and possibly integrating additional contextual information during training.

Looking forward, there are several promising avenues for future work that can extend the contributions of this paper. First, optimizing the training process to reduce computational overhead and improve scalability is a crucial next step. This could involve developing more efficient attention mechanisms or exploring alternative architectural designs that balance performance with resource consumption. Second, enhancing the model's capability to interpret complex scenes and abstract concepts remains a significant challenge. Future research could focus on integrating external knowledge sources, such as knowledge graphs, or leveraging advanced pre-trained models with larger and more diverse datasets to enrich the model's contextual understanding. Additionally, exploring ways to handle varying input complexities and ensuring robustness across different real-world scenarios will be important for practical deployment. Our study demonstrates the effectiveness of combining ranking-based hybrid training with multimodal fusion, advancing the state-of-the-art in VQA. By emphasizing the importance of attention mechanisms and sophisticated feature integration, our research lays a solid groundwork for the development of more advanced and accurate VQA systems. Future efforts should build on these insights to further enhance the performance and applicability of VQA models.

In conclusion, our study presents a significant step forward in the field of Visual Question Answering by introducing a model that combines ranking-based training and multimodal fusion to achieve high accuracy and robustness. While there are areas for improvement, the advancements made in this work lay the groundwork for future research aimed at further enhancing the capabilities of VQA systems. By continuing to explore and refine these techniques, we can move closer to developing models that not only understand and answer questions about images with high accuracy but also do so efficiently and with a deeper understanding of the visual content.

## Acknowledgments

Not applicable.